\RequirePackage{silence}
\documentclass[runningheads]{llncs}

\usepackage{xcolor}
\usepackage{graphicx}
\graphicspath{{/}{fig/}}
\usepackage{cite}
\usepackage{tikz}
\usepackage{array}
\usepackage{textcomp}
\usepackage{xcolor}
\usepackage{multirow}
\usepackage{booktabs}

\usepackage[style=base]{caption}
\usepackage{subcaption}

\usepackage{mathtools}
\usepackage{float}
\usepackage{hyperref}

\usepackage{pgfplots}
\pgfplotsset{compat=1.7}
\usepgfplotslibrary{groupplots}

\usepackage{multirow,tabularx}

\usepackage{tikz}

\newlength\figureheight
\newlength\figurewidth
\begin{document}
\title{Autocalibration of a Mobile UWB Localization System for Ad-Hoc Multi-Robot Deployments in GNSS-Denied Environments}
\author{
    Carmen Mart\'{i}nez Almansa\inst{1} \and
    Wang Shule\inst{1} \and \\[+4.2pt]
    Jorge Pe\~{n}a Queralta\inst{1} \and
    Tomi Westerlund\inst{1}
}%
\authorrunning{Carmen Mart\'{i}nez Almansa et al.}
\titlerunning{Autocalibration of a Mobile UWB Localization System}
\institute{
        \href{https://tiers.utu.fi}{Turku Intelligent Embedded and Robotic Systems} \\
        University of Turku, Finland \\
        \email{\{camart, shwang, jopequ, tovewe\}@utu.fi} \\
        \url{https://tiers.utu.fi}
    }
\maketitle
%
%
%
\begin{abstract}
    Ultra-wideband (UWB) wireless technology has seen an increased penetration in the robotics field as a robust localization method in recent years. UWB enables high accuracy distance estimation from time-of-flight measurements of wireless signals, even in non-line-of-sight measurements. UWB-based localization systems have been utilized in various types of GNSS-denied environments for ground or aerial autonomous robots. However, most of the existing solutions rely on a fixed and well-calibrated set of UWB nodes, or anchors, to estimate accurately the position of other mobile nodes, or tags, through multilateration. This limits the applicability of such systems for dynamic and ad-hoc deployments, such as post-disaster scenarios where the UWB anchors could be mounted on mobile robots to aid the navigation of UAVs or other robots. We introduce a collaborative algorithm for online autocalibration of anchor positions, enabling not only ad-hoc deployments but also movable anchors, based on Decawave's DWM1001 UWB module. Compared to the built-in autocalibration process from Decawave, we drastically reduce the amount of calibration time and increase the accuracy at the same time. We provide both experimental measurements and simulation results to demonstrate the usability of this algorithm.
\end{abstract}


\section{Introduction}
\label{sec:intro}


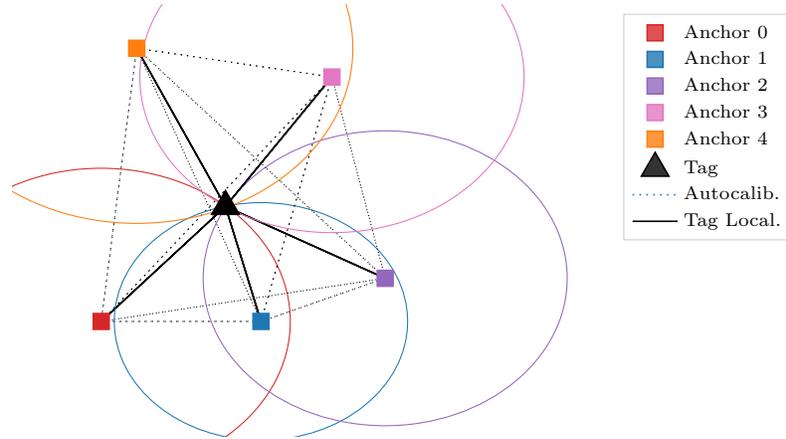
\begin{figure}
    \centering
    \setlength{\figureheight}{0.6\textwidth}
    \setlength{\figurewidth}{\textwidth}
    \scriptsize{
\begin{tikzpicture}

\definecolor{color0}{rgb}{0.83921568627451,0.152941176470588,0.156862745098039}
\definecolor{color1}{rgb}{0.12156862745098,0.466666666666667,0.705882352941177}
\definecolor{color2}{rgb}{0.580392156862745,0.403921568627451,0.741176470588235}
\definecolor{color3}{rgb}{0.890196078431372,0.466666666666667,0.76078431372549}
\definecolor{color4}{rgb}{1,0.498039215686275,0.0549019607843137}

\begin{axis}[
    name=ax1,
    height=\figureheight,
    width=\figurewidth,
    axis line style={white},
    legend cell align={left},
    legend style={fill opacity=0.8, draw opacity=1, text opacity=1, draw=white!80!black},
    tick align=outside,
    tick pos=left,
    x grid style={white},
    xmin=-3, xmax=42,
    xtick style={color=black},
    y grid style={white},
    ymin=-5, ymax=25,
    ytick style={color=black},
    axis line style={draw=none},
    tick style={draw=none},
    xtick={5,10,15},
    ytick={5,10,15},
    xticklabels={,,},
    yticklabels={,,}
]
\draw[draw=color0] (axis cs:2,3) circle (10.6540925797477);
\path [draw=black, fill=color1, semithick]
(axis cs:9.00296327073541,11.0033865951262)
--(axis cs:9.00112886504206,10.9990122430882)
--(axis cs:9.00037628834735,10.9996707476961)
--(axis cs:2.00037628834735,2.99967074769607)
--(axis cs:1.99962371165265,3.00032925230393)
--(axis cs:8.99962371165265,11.0003292523039)
--(axis cs:8.99887113495794,11.0009877569118)
--cycle;

\path [draw=black, fill=color1, dotted]
(axis cs:11.0045,3)
--(axis cs:11,2.9985)
--(axis cs:11,2.9995)
--(axis cs:2,2.9995)
--(axis cs:2,3.0005)
--(axis cs:11,3.0005)
--(axis cs:11,3.0015)
--cycle;

\path [draw=black, fill=color1, dotted]
(axis cs:18.0044229248412,6.00082929840773)
--(axis cs:18.0002764328026,5.9985256917196)
--(axis cs:18.0000921442675,5.99950856390653)
--(axis cs:2.00009214426752,2.99950856390653)
--(axis cs:1.99990785573247,3.00049143609347)
--(axis cs:17.9999078557325,6.00049143609347)
--(axis cs:17.9997235671974,6.0014743082804)
--cycle;

\path [draw=black, fill=color1, dotted]
(axis cs:15.0027335254841,20.0035746102484)
--(axis cs:15.0011915367495,19.9990888248386)
--(axis cs:15.0003971789165,19.9996962749462)
--(axis cs:2.00039717891649,2.99969627494621)
--(axis cs:1.99960282108351,3.00030372505379)
--(axis cs:14.9996028210835,20.0003037250538)
--(axis cs:14.9988084632505,20.0009111751614)
--cycle;

\path [draw=black, fill=color1, dotted]
(axis cs:4.00047108153033,22.0044752745381)
--(axis cs:4.00149175817938,21.9998429728232)
--(axis cs:4.00049725272646,21.9999476576077)
--(axis cs:2.00049725272646,2.99994765760774)
--(axis cs:1.99950274727354,3.00005234239226)
--(axis cs:3.99950274727354,22.0000523423923)
--(axis cs:3.99850824182062,22.0001570271768)
--cycle;

\draw[draw=color1] (axis cs:11,3) circle (8.26315572214983);
\path [draw=black, fill=color1, semithick]
(axis cs:8.99890858968734,11.0043656412507)
--(axis cs:9.00145521375022,11.0003638034376)
--(axis cs:9.00048507125007,11.0001212678125)
--(axis cs:11.0004850712501,3.00012126781252)
--(axis cs:10.9995149287499,2.99987873218748)
--(axis cs:8.99951492874993,10.9998787321875)
--(axis cs:8.99854478624978,10.9996361965624)
--cycle;

\path [draw=black, fill=color1, dotted]
(axis cs:18.0041361526351,6.00177263684361)
--(axis cs:18.0005908789479,5.99862128245497)
--(axis cs:18.0001969596493,5.99954042748499)
--(axis cs:11.0001969596493,2.99954042748499)
--(axis cs:10.9998030403507,3.00045957251501)
--(axis cs:17.9998030403507,6.00045957251501)
--(axis cs:17.9994091210521,6.00137871754503)
--cycle;

\path [draw=black, fill=color1, dotted]
(axis cs:15.0010306770018,20.0043803772575)
--(axis cs:15.0014601257525,19.9996564409994)
--(axis cs:15.0004867085842,19.9998854803331)
--(axis cs:11.0004867085842,2.99988548033314)
--(axis cs:10.9995132914158,3.00011451966686)
--(axis cs:14.9995132914158,20.0001145196669)
--(axis cs:14.9985398742475,20.0003435590006)
--cycle;

\path [draw=black, fill=color1, dotted]
(axis cs:3.99844432588528,22.0042225440257)
--(axis cs:4.00140751467523,22.0005185580382)
--(axis cs:4.00046917155841,22.0001728526794)
--(axis cs:11.0004691715584,3.00017285267941)
--(axis cs:10.9995308284416,2.99982714732059)
--(axis cs:3.99953082844159,21.9998271473206)
--(axis cs:3.99859248532477,21.9994814419618)
--cycle;

\draw[draw=color2] (axis cs:18,6) circle (10.2543355694359);
\path [draw=black, fill=color1, semithick]
(axis cs:8.99606629225745,11.0021853931903)
--(axis cs:9.00072846439677,11.0013112359142)
--(axis cs:9.00024282146559,11.0004370786381)
--(axis cs:18.0002428214656,6.00043707863806)
--(axis cs:17.9997571785344,5.99956292136194)
--(axis cs:8.99975717853441,10.9995629213619)
--(axis cs:8.99927153560323,10.9986887640858)
--cycle;

\path [draw=black, fill=color1, dotted]
(axis cs:14.9990571191007,20.0044001108633)
--(axis cs:15.0014667036211,20.0003142936331)
--(axis cs:15.000488901207,20.0001047645444)
--(axis cs:18.000488901207,6.00010476454437)
--(axis cs:17.999511098793,5.99989523545563)
--(axis cs:14.999511098793,19.9998952354556)
--(axis cs:14.9985332963789,19.9996857063669)
--cycle;

\path [draw=black, fill=color1, dotted]
(axis cs:3.99703672926459,22.0033865951262)
--(axis cs:4.00112886504206,22.0009877569118)
--(axis cs:4.00037628834735,22.0003292523039)
--(axis cs:18.0003762883474,6.00032925230393)
--(axis cs:17.9996237116526,5.99967074769607)
--(axis cs:3.99962371165265,21.9996707476961)
--(axis cs:3.99887113495794,21.9990122430882)
--cycle;

\draw[draw=color3] (axis cs:15,20) circle (10.8243983565116);
\path [draw=black, fill=color1, semithick]
(axis cs:9,11)
--(axis cs:8.99875192455849,11.0008320502943)
--(axis cs:8.99958397485283,11.0002773500981)
--(axis cs:15,20)
--(axis cs:15.0004160251472,19.9997226499019)
--(axis cs:9.00041602514717,10.9997226499019)
--(axis cs:9.00124807544151,10.9991679497057)
--cycle;

\path [draw=black, fill=color1, dotted]
(axis cs:3.99557258540455,22.0008049844719)
--(axis cs:4.0002683281573,22.0014758048652)
--(axis cs:4.0000894427191,22.0004919349551)
--(axis cs:15.0000894427191,20.0004919349551)
--(axis cs:14.9999105572809,19.9995080650449)
--(axis cs:3.9999105572809,21.9995080650449)
--(axis cs:3.9997316718427,21.9985241951348)
--cycle;

\draw[draw=color4] (axis cs:4,22) circle (12.1783843704583);
\path [draw=black, fill=color1, semithick]
(axis cs:9.00186211324936,10.9959033508514)
--(axis cs:8.99863445028381,10.9993792955835)
--(axis cs:8.99954481676127,10.9997930985278)
--(axis cs:3.99954481676127,21.9997930985278)
--(axis cs:4.00045518323873,22.0002069014722)
--(axis cs:9.00045518323873,11.0002069014722)
--(axis cs:9.00136554971619,11.0006207044165)
--cycle;

\addplot [semithick, color0, mark=square*, mark size=3, mark options={solid}, only marks]
table {%
2 3
};
\addlegendentry{Anchor 0}
\addplot [semithick, color1, mark=square*, mark size=3, mark options={solid}, only marks]
table {%
11 3
};
\addlegendentry{Anchor 1}
\addplot [semithick, color2, mark=square*, mark size=3, mark options={solid}, only marks]
table {%
18 6
};
\addlegendentry{Anchor 2}
\addplot [semithick, color3, mark=square*, mark size=3, mark options={solid}, only marks]
table {%
15 20
};
\addlegendentry{Anchor 3}
\addplot [semithick, color4, mark=square*, mark size=3, mark options={solid}, only marks]
table {%
4 22
};
\addlegendentry{Anchor 4}
\addplot [semithick, black, mark=triangle*, mark size=6, mark options={solid}, only marks]
table {%
9 11
};
\addlegendentry{Tag}
\addplot [semithick, color1, dotted]
table {%
0 0
0.001 0.001
};
\addlegendentry{Autocalib.}
\addplot [semithick, black, semithick]
table {%
0 0
0.001 0.001
};
\addlegendentry{Tag Local.}
\end{axis}

\end{tikzpicture}}
    \caption{UWB localization and autocalibration concept. The circles are defined by the UWB-measured range between each of the anchors and the tag. The dotted lines represent the inter-anchor measurements taken during the autocalibration process. The number of tags and anchors is arbitrary, and only one tag is shown here for illustrative purposes.}
    \label{fig:concept}
\end{figure}


The utilization of UWB radios for both localization and short-range data transmission started to gain momentum after the unlicensed usage legalization in 2002~\cite{fcc2002report}, and the IEEE standards released in 2007~\cite{karapistoli2010overview}. Nonetheless, only in recent years UWB-based localization systems have seen wider adoption in the robotics domain, owing to their high accuracy, and often as a replacement to GNSS sensors in GNSS-denied environments~\cite{macoir2019uwb}. UWB-based systems are now being utilized for communication and localization~\cite{nguyen2019integrated}, or as short-range radar systems for mapping or navigation, among other applications~\cite{taylor2018ultra}.

UWB-based localization systems provide an inexpensive alternative to high-accuracy motion capture systems for navigation in application scenarios where a localization accuracy of the order of tens of centimeters is sufficient~\cite{furtado2019comparative}. In GNSS-denied environments, UWB-based localization systems can provide a robust alternative to visual odometry methods~\cite{qingqing2019monocular}, or other methods that rely only on information acquired onboard mobile agents, such as lidar odometry~\cite{sarker2019offloading}, which present challenges in long-term autonomy. Therefore, UWB-based localization systems enable longer operations and tighter control over the behavior of mobile robots. Moreover, accurate relative localization in multi-robot systems can aid information control algorithms, such as those where only relative position estimation is needed~\cite{guo2019ultra, queralta2019indexfree, queralta2019progressive}, or collaborative tasks requiring multi-source sensor fusion~\cite{queralta2020blockchain}, such as cooperative mapping~\cite{queralta2019collaborative} or docking of unmanned aerial vehicles (UAVs) on mobile platforms~\cite{nguyen2019integrated}.

One of the main limitations of UWB-based localization systems, which they share with many other wireless localization systems based on active beacons, is that they require a predefined set of beacons to be located in known positions in the operational environment~\cite{zafari2019survey}. In UWB systems, these fixed radio nodes are often called anchors, while mobile nodes are called tags. Fixed anchors are required because only ranging information can be extracted from UWB signals. From a set of at least three anchor-tag distance measurements, the position of a tag can be calculated from the anchors' positions utilizing multilateration methods~\cite{queralta2020uwb}.

Current systems, which mainly rely on a fixed set of anchors as a reference, require accurate calibration of the anchor positions, this significantly limiting their applicability. Motivated by this, we have developed an automatic calibration method that allows these anchors to be mobile and hence to be used in dynamic localization systems. The typical procedure to estimate the position of a mobile tag based on the position of fixed anchors is depicted in Fig.~\ref{fig:concept}, where the radius of each circle is defined by the distance to the tag estimated through UWB ranging. The tag can locate itself by estimating the individual distances to each of the anchors (solid line), while inter-anchor distances (dotted lines) can be utilized by the anchors themselves to calibrate their positions.

In summary, our main objective is the design and development of a mobile UWB-based localization system that can be utilized for localization in multi-robot systems in GNSS-denied environments. This paper presents initial results in this direction. The DWM1001 UWB transceiver from Decawave has been utilized and we have developed an autocalibration as part of wider UWB experiments reported in~\cite{queralta2020uwb}. The code is made publicly available in our GitHub repository\footnote{\href{https://github.com/tiers/uwb_drone_dataset}{TIERS UWB Dataset: https://github.com/tiers/uwb\_drone\_dataset}}, where we have released an initial version of the autocalibration firmware for Decawave's DWM1001 development board. We utilize UWB accuracy measurements from our experiments to simulate the performance of a mobile UWB-based localization system. This paper, therefore, focuses on the results of those simulations to assess the viability and usability of the proposed system.

The remainder of this paper is organized as follows. In Section 2, we review related works regarding the autocalibration of UWB localization systems and provide a broad overview of their potential applications. Section 3 then introduces the details about the UWB calibration and localization process, with initial results reported in Section 4. Finally, Section 5 concludes the work and outlines future research directions.

\section{Related work}
\label{sec:related}

In this section, we first review existing autocalibration methods for UWB-based localization systems. Then, we analyze in more detail the autocalibration method included in the Decawave's firmware, as well as its requirements and drawbacks.

An early approach to automatic calibration of UWB radios in mobile robots localization systems was proposed by K. C. Cheok \textit{et al.}~\cite{5671780}. The algorithm proposed by the authors is capable of determining the coordinates of four anchors from UWB measurements estimating the distance between each pair of anchors. The algorithm relies on the following assumptions to calculate the anchor positions: there must exist a known order of the four anchors such as anchor 0 defines the origin of coordinates; anchor 0 and 1 define the positive x-axis direction; and the plane x-y is defined by the first three anchors. 

Another autocalibration UWB-based multi-robot localization system presented by M. Hamer \textit{et al.} is stricter in terms of assumptions~\cite{8344407}. In addition to the aforementioned conditions, in this second system it is also assumed that anchor 2 lies on the positive y-direction, anchor 3 on the positive z-axis and all anchors are at fixed positions. Moreover, the system relies on clock synchronization, since the localization is based on time difference of arrival (TDoA).

Several other works have presented on-board localization systems based on UWB technology for either one target~\cite{8511568, 7759054}, or multiple targets~\cite{8814678}. In these papers, the anchors are situated on a mobile platform. The relative position of the tag, which is mounted on the target robot or person, is estimated from the distances between itself and the anchors.

Regarding Decawave's UWB modules, a built-in calibration system is available through their mobile application as part of Decawave's real-time localization system (DRTLS). This process. called auto-positioning, can be utilized with a minimum a priori knowledge of the anchor positions: it requires the anchors (up to four) to be arranged in a rectangular shape, at an equal or similar height, and in counter-clockwise order. In addition to this, we have found the calculation time of this algorithm to be around 40\,s and the error above 1\,m in deployments where the inter-anchor distance was less than 20\,m. These characteristics make the algorithm overly slow and inaccurate to be suitable for mobile settings. The lack of accuracy is warned in the app itself, where it is recommended to measure and introduce the anchor positions manually since the autocalibration feature makes the positioning less precise. Decawave devices are some of the most widely used UWB ranging modules~\cite{pan2019dwm1000}, and thus there is an evident need for faster and more accurate autocalibration methods to enable faster ad-hoc and even mobile deployments.

\section{Autocalibrating a Mobile UWB Localization System}

In this section, we first describe how distance can be estimated from the time of flight of a UWB signal, and then introduce our proposed autocalibration method for the anchors.

\begin{figure}[ht!]
    \centering
    \includegraphics[width=.6\textwidth]{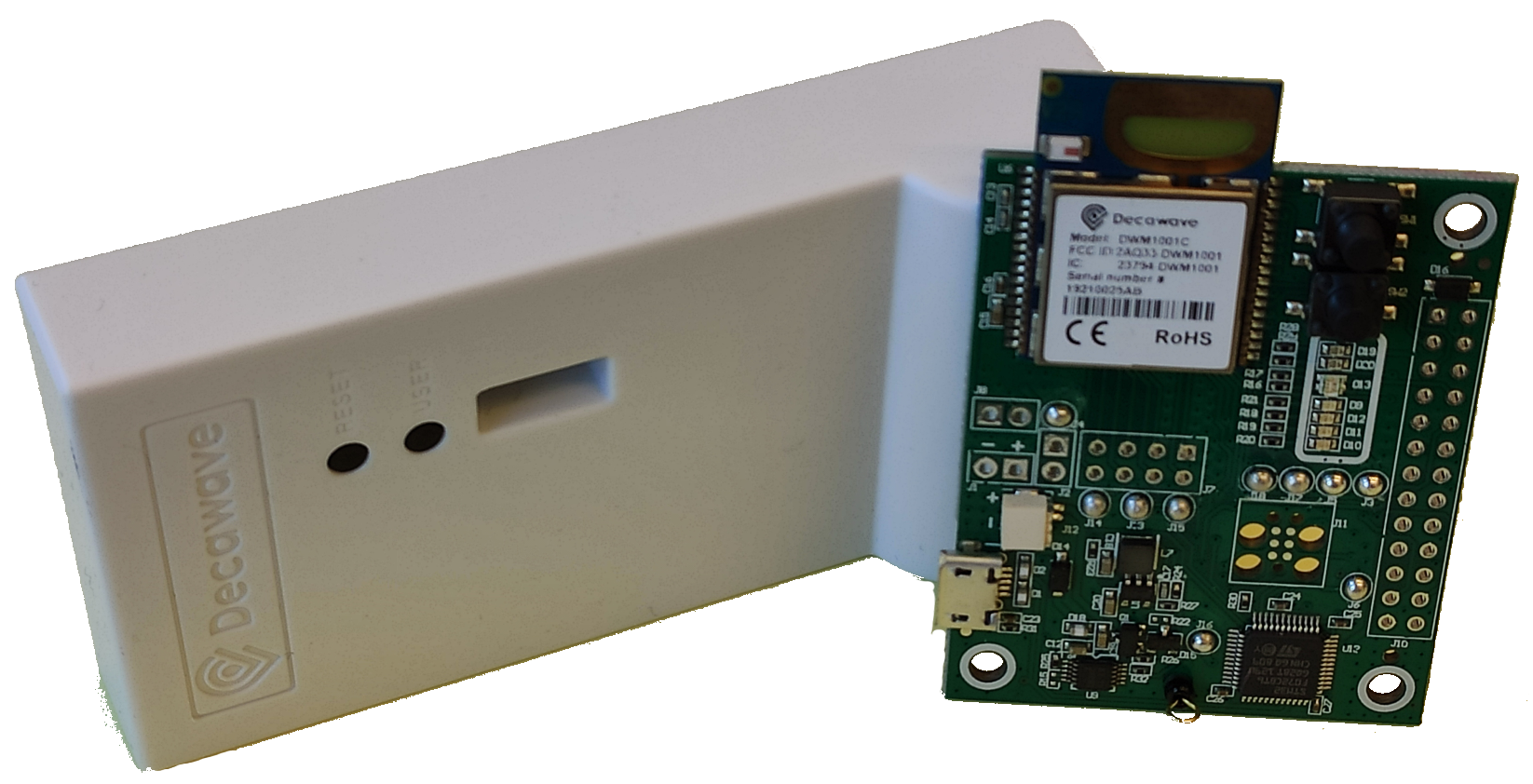}
    \caption{Decawave's DWM1001 Development Board, which has been utilized in the experiments}
    \label{fig:dwm1001}
\end{figure}

\subsection{UWB Ranging}

The two main methods for UWB ranging measurements, also applicable to other wireless ranging technologies, are time of flight (ToF) and time difference of arrival (TDoA).

ToF is a method for estimating the distance between an emitter and a receiver node multiplying the time of flight of the signal between a single pair of transceivers, usually an anchor and a tag, by the speed of light in air~\cite{mazraani2017experimental}. It's a two-way ranging (TWR) technology, requiring transmissions in both directions. In single-side TWR (SS-TWR), a transmitter, or initiator, sends a poll message which then receiving node replies to. By measuring the total time until it obtains a response, the initiator can then estimate the distance that separates it from the node that replied to the message. In this situation, the antenna delays and the fixed time required to process the poll message and send the response at the receiving node must be known and taken into account when estimating the distance. Double-side TWR (DS-TWR) eliminates the need for calibration by adding an additional response, or final message.

TDoA is another widely-used method for locating a mobile node by detecting the time difference of arrival (TDoA) of the same wireless signal received at multiple interconnected anchors~\cite{wei2018joint}. In this algorithm, the anchors need to be synchronized, and the hyperbolic branch is drawn for each anchor pair from the difference between the reception time of the main anchor and other anchors~\cite{mazraani2017experimental}. The point where all the hyperbolic intersections occur is taken as the approximate location of the tag. TDoA ranging is also called hyperbolic ranging.

\subsection{Autocalibration of Anchors}

The aim of our work is to develop a UWB-based localization system with built-in autocalibration, which could be used for the localization of multi-robot systems in dynamic scenarios. Our customized autocalibration method relies on a series of assumptions for the first measurement, in order to localize the system in the space. These initial assumptions are similar to those in the related works described in the previous section:
\begin{itemize}
    \item The first anchor (Anchor 0) is situated at the origin of coordinates. 
    \item The direction from Anchor 0 to Anchor 1 defines the positive x-axis.
    \item All other anchors lie in the half-plane with positive y-coordinate.
\end{itemize}

Based on these assumptions, the initial calibration step estimates the position of each of the anchors based on the measured distances to the first two anchors defining the origin of coordinates and the positive x-axis direction. Then, the position of all anchors is adjusted by minimizing the error between the inter-anchor distances and the UWB ranging measurements with a least squares estimator (LSE). After the initial calibration step, the only assumption we make is that the position anchor 0 defines the origin. The reason behind this relaxed conditions regarding the x and y axis is that our experiments have shown that the rotational error is negligible. This implies more flexible conditions than in previous works~\cite{8344407} and~\cite{5671780}.

In our autocalibration process, every anchor behaves as initiator and responder in turns. The anchor that defines the origin is the first initiator. The process is initiated by a start command sent to the corresponding anchor through the UART interface. This first initiator, henceforth referred to as Anchor 0, calculates the distances to each of the other anchors. The distances are estimated based on the time of flight (TOF) using SS-TWR. The communication is done in pairs, only after receiving the distance measurement from one responder and broadcasting it, the initiator will start communication with the following one. 


Once the initiator has gathered the distance values to every other anchor, it will send a message to the following one, according to the counter-clockwise order established, and will change its mode to responder. The recipient of the message will become initiator and start the cycle again. When the last anchor in the network finishes its measurements, it will send the message to the Anchor 0, which will become initiator again, and await the next start trigger. Calibrations should occur periodically whenever the inter-calibration positioning error at the anchors exceeds a certain error threshold. The inter-calibration positioning can be done with other on-board methods, such as visual or lidar odometry.

\begin{table}
    \centering
    \caption{Latency of the Autopositioning method from Decawave's DRTLS compared to our self-calibration method for anchors.}
    \small
    \begin{tabular}{@{}ll@{}}
        \toprule
        & \hspace{0.42cm}\textbf{Latency}  \\ 
        \midrule
        \textbf{RTLS Autopositioning}       & \hspace{0.42cm}$40\,s\:\pm5\,s$    \\
        \textbf{Custom Calibration (x50)}   & \hspace{0.42cm}$2.5\,s\pm0.1\,s$    \\
        \textbf{Custom Calibration (x5)}    & \hspace{0.42cm}$0.9\,s\pm0.05\,s$   \\
        \bottomrule
    \end{tabular}
    \label{tab:latencies}
\end{table}

\begin{table}
    \centering
    \caption{Accuracy of the Autopositioning method from Decawave's DRTLS compared to our self-calibration method for anchors.}
    \small
    \begin{tabular}{@{}lrrrr@{}}
        \toprule
        \textbf{Covered}\hspace{0.42cm} & \multicolumn{2}{c}{\textbf{RTLS Autopositioning}} & \multicolumn{2}{c}{\textbf{Our Autocalibration}} \\ 
        \textbf{\:\:\:Area} & \hspace{0.42cm}Min. Err. & Max. Err.\hspace{0.42cm} & \hspace{0.42cm}Min. Err. & Max. Err.\hspace{0.42cm} \\
        \midrule
        \textbf{$1\,m^2$}       &  9\,cm  & 52\,cm\hspace{0.42cm} & 4\,cm & 39\,cm\hspace{0.42cm} \\
        \textbf{$9\,m^2$}       &  4\,cm  & 28\,cm\hspace{0.42cm} & 5\,cm & 24\,cm\hspace{0.42cm} \\
        \textbf{$144\,m^2$}     &  163\,cm  & 219\,cm\hspace{0.42cm} & 135\,cm & 182\,cm\hspace{0.42cm} \\
        \bottomrule
    \end{tabular}
    \label{tab:autopositioning}
\end{table}

This autocalibration process has been implemented in C and the firmware for Decawave's DWM1001 Development board, illustrated in Fig.~\ref{fig:dwm1001}, has been made publicly available in Github. Table~\ref{tab:autopositioning} shows the difference in calibration accuracy between our firmware and Decawave's DRTLS autopositioning system, the latter being a process that is triggered through the DRLTS mobile application. In our implementation, every time the autocalibration occurs, multiple measurements are taken and the average and standard deviation are shared with all other anchors to estimate each other's positions. In Table~\ref{tab:latencies}, we show the latency when we take 5 or 50 measurements for each pair of anchors.

\begin{figure}
    \centering
    \setlength{\figureheight}{0.42\textwidth}
    \setlength{\figurewidth}{0.95\textwidth} 
    \scriptsize{
\begin{tikzpicture}

\definecolor{color0}{rgb}{0.580392156862745,0.403921568627451,0.741176470588235}

\begin{axis}[
    name=ax1,
    height=\figureheight,
    width=\figurewidth,
    axis background/.style={fill=white!92!black},
    axis line style={white},
    legend cell align={left},
    legend style={fill opacity=0.8, draw opacity=1, text opacity=1, at={(0.03,0.97)}, anchor=north west, draw=white!80!black},
    tick align=outside,
    tick pos=left,
    x grid style={white},
    xlabel={Real Distance (m)},
    xmajorgrids,
    xmin=-0.575, xmax=23.075,
    xtick style={color=black},
    y grid style={white},
    ylabel={Measured Distance (m)},
    ymajorgrids,
    ymin=-0.308553261952495, ymax=23.6396185010024,
    ytick style={color=black}
]
\addplot [semithick, black]
table {%
0.5 0.865413404587226
0.6 0.966276901491704
0.7 1.06714039839618
0.8 1.16800389530066
0.9 1.26886739220514
1 1.36973088910961
1.2 1.57145788291857
1.4 1.77318487672752
1.6 1.97491187053648
1.8 2.17663886434543
2 2.37836585815439
2.4 2.7818198457723
2.8 3.18527383339021
3.2 3.58872782100812
3.6 3.99218180862603
4 4.39563579624394
4.8 5.20254377147976
5.2 5.60599775909767
5.6 6.00945174671558
6 6.41290573433349
6.4 6.8163597219514
6.8 7.21981370956931
7.2 7.62326769718722
7.6 8.02672168480513
8 8.43017567242305
9 9.43881064146782
10 10.4474456105126
11 11.4560805795574
12 12.4647155486021
13 13.4733505176469
14 14.4819854866917
15 15.4906204557365
16 16.4992554247813
17 17.507890393826
18 18.5165253628708
19 19.5251603319156
20 20.5337953009604
21 21.5424302700051
22 22.5510652390499
};
\addlegendentry{Linear fit}
\addplot [semithick, color0, mark=square*, mark size=1.23, mark options={solid}, forget plot]
table {%
0.5 0.78
};
\addplot [semithick, color0, mark=square*, mark size=1.23, mark options={solid}, forget plot]
table {%
0.6 0.86
};
\addplot [semithick, color0, mark=square*, mark size=1.23, mark options={solid}, forget plot]
table {%
0.7 1.09
};
\addplot [semithick, color0, mark=square*, mark size=1.23, mark options={solid}, forget plot]
table {%
0.8 1.06
};
\addplot [semithick, color0, mark=square*, mark size=1.23, mark options={solid}, forget plot]
table {%
0.9 1.2
};
\addplot [semithick, color0, mark=square*, mark size=1.23, mark options={solid}, forget plot]
table {%
1 1.35
};
\addplot [semithick, color0, mark=square*, mark size=1.23, mark options={solid}, forget plot]
table {%
1.2 1.58
};
\addplot [semithick, color0, mark=square*, mark size=1.23, mark options={solid}, forget plot]
table {%
1.4 1.82
};
\addplot [semithick, color0, mark=square*, mark size=1.23, mark options={solid}, forget plot]
table {%
1.6 1.91
};
\addplot [semithick, color0, mark=square*, mark size=1.23, mark options={solid}, forget plot]
table {%
1.8 2.1
};
\addplot [semithick, color0, mark=square*, mark size=1.23, mark options={solid}, forget plot]
table {%
2 2.34
};
\addplot [semithick, color0, mark=square*, mark size=1.23, mark options={solid}, forget plot]
table {%
2.4 2.76
};
\addplot [semithick, color0, mark=square*, mark size=1.23, mark options={solid}, forget plot]
table {%
2.8 3.18
};
\addplot [semithick, color0, mark=square*, mark size=1.23, mark options={solid}, forget plot]
table {%
3.2 3.63
};
\addplot [semithick, color0, mark=square*, mark size=1.23, mark options={solid}, forget plot]
table {%
3.6 4.02
};
\addplot [semithick, color0, mark=square*, mark size=1.23, mark options={solid}, forget plot]
table {%
4 4.45
};
\addplot [semithick, color0, mark=square*, mark size=1.23, mark options={solid}, forget plot]
table {%
4.8 5.3
};
\addplot [semithick, color0, mark=square*, mark size=1.23, mark options={solid}, forget plot]
table {%
5.2 5.68
};
\addplot [semithick, color0, mark=square*, mark size=1.23, mark options={solid}, forget plot]
table {%
5.6 6.08
};
\addplot [semithick, color0, mark=square*, mark size=1.23, mark options={solid}, forget plot]
table {%
6 6.47
};
\addplot [semithick, color0, mark=square*, mark size=1.23, mark options={solid}, forget plot]
table {%
6.4 6.78
};
\addplot [semithick, color0, mark=square*, mark size=1.23, mark options={solid}, forget plot]
table {%
6.8 7.26
};
\addplot [semithick, color0, mark=square*, mark size=1.23, mark options={solid}, forget plot]
table {%
7.2 7.67
};
\addplot [semithick, color0, mark=square*, mark size=1.23, mark options={solid}, forget plot]
table {%
7.6 8.12
};
\addplot [semithick, color0, mark=square*, mark size=1.23, mark options={solid}, forget plot]
table {%
8 8.44
};
\addplot [semithick, color0, mark=square*, mark size=1.23, mark options={solid}, forget plot]
table {%
9 9.53
};
\addplot [semithick, color0, mark=square*, mark size=1.23, mark options={solid}, forget plot]
table {%
10 10.43
};
\addplot [semithick, color0, mark=square*, mark size=1.23, mark options={solid}, forget plot]
table {%
11 11.47
};
\addplot [semithick, color0, mark=square*, mark size=1.23, mark options={solid}, forget plot]
table {%
12 12.44
};
\addplot [semithick, color0, mark=square*, mark size=1.23, mark options={solid}, forget plot]
table {%
13 13.49
};
\addplot [semithick, color0, mark=square*, mark size=1.23, mark options={solid}, forget plot]
table {%
14 14.55
};
\addplot [semithick, color0, mark=square*, mark size=1.23, mark options={solid}, forget plot]
table {%
15 15.53
};
\addplot [semithick, color0, mark=square*, mark size=1.23, mark options={solid}, forget plot]
table {%
16 16.53
};
\addplot [semithick, color0, mark=square*, mark size=1.23, mark options={solid}, forget plot]
table {%
17 17.48
};
\addplot [semithick, color0, mark=square*, mark size=1.23, mark options={solid}, forget plot]
table {%
18 18.5
};
\addplot [semithick, color0, mark=square*, mark size=1.23, mark options={solid}, forget plot]
table {%
19 19.42
};
\addplot [semithick, color0, mark=square*, mark size=1.23, mark options={solid}, forget plot]
table {%
20 20.48
};
\addplot [semithick, color0, mark=square*, mark size=1.23, mark options={solid}, forget plot]
table {%
21 21.47
};
\addplot [semithick, color0, mark=square*, mark size=1.23, mark options={solid}, forget plot]
table {%
22 22.55
};
\addplot [color0, mark=square*, mark size=1.23, mark options={solid}]
table {%
22 22.55
};
\addlegendentry{Raw data}

\coordinate (c1) at (axis cs:0,0);
\coordinate (c2) at (axis cs:2,2.5);
\draw [dashed, gray] (c1) rectangle (axis cs:2,2.5);

\end{axis}

\begin{axis}[
    name=ax2,
    at={($(ax1.south west)+(0.14\textwidth, -0.8\figureheight)$)},
    height=0.8\figureheight,
    width=(0.72\figurewidth),
    axis background/.style={fill=white!92!black},
    axis line style={white},
    legend cell align={left},
    legend style={fill opacity=0.8, draw opacity=1, text opacity=1, at={(0.03,0.97)}, anchor=north west, draw=white!80!black},
    tick align=outside,
    tick pos=left,
    x grid style={white},
    xlabel={Real Distance (m)},
    xmajorgrids,
    xmin=0.425, xmax=2.075,
    xtick style={color=black},
    xtick={0.4,0.6,0.8,1,1.2,1.4,1.6,1.8,2,2.2},
    xticklabels={0.4,0.6,0.8,1.0,1.2,1.4,1.6,1.8,2.0,2.2},
    y grid style={white},
    ylabel={Meas. Distance (m)},
    ymajorgrids,
    ymin=0.700081707092281, ymax=2.45828415106211,
    ytick style={color=black},
    ytick={0.6,0.8,1,1.2,1.4,1.6,1.8,2,2.2,2.4,2.6},
    yticklabels={0.2,0.4,0.6,0.8,1.0,1.2,1.4,1.6,1.8,2.0,2.2}
]
\addplot [semithick, black]
table {%
0.5 0.865413404587226
0.6 0.966276901491704
0.7 1.06714039839618
0.8 1.16800389530066
0.9 1.26886739220514
1 1.36973088910961
1.2 1.57145788291857
1.4 1.77318487672752
1.6 1.97491187053648
1.8 2.17663886434543
2 2.37836585815439
};
\addlegendentry{Linear fit}
\addplot [semithick, color0, mark=square*, mark size=1.42, mark options={solid}, forget plot]
table {%
0.5 0.78
};
\addplot [semithick, color0, mark=square*, mark size=1.42, mark options={solid}, forget plot]
table {%
0.6 0.86
};
\addplot [semithick, color0, mark=square*, mark size=1.42, mark options={solid}, forget plot]
table {%
0.7 1.09
};
\addplot [semithick, color0, mark=square*, mark size=1.42, mark options={solid}, forget plot]
table {%
0.8 1.06
};
\addplot [semithick, color0, mark=square*, mark size=1.42, mark options={solid}, forget plot]
table {%
0.9 1.2
};
\addplot [semithick, color0, mark=square*, mark size=1.42, mark options={solid}, forget plot]
table {%
1 1.35
};
\addplot [semithick, color0, mark=square*, mark size=1.42, mark options={solid}, forget plot]
table {%
1.2 1.58
};
\addplot [semithick, color0, mark=square*, mark size=1.42, mark options={solid}, forget plot]
table {%
1.4 1.82
};
\addplot [semithick, color0, mark=square*, mark size=1.42, mark options={solid}, forget plot]
table {%
1.6 1.91
};
\addplot [semithick, color0, mark=square*, mark size=1.42, mark options={solid}, forget plot]
table {%
1.8 2.1
};
\addplot [semithick, color0, mark=square*, mark size=1.42, mark options={solid}]
table {%
2 2.34
};
\addlegendentry{Raw data}
\end{axis}

\begin{axis}[
    at={(ax2.south west)},
    axis line style={draw=none},
    tick style={draw=none},
    xmin=0,xmax=42,
    ymin=0,ymax=42,
    xtick={5,10,15},
    ytick={5,10,15},
    xticklabels={,,},
    yticklabels={,,}
]

\end{axis}

\draw [dashed, gray] (c1) -- (ax2.south west);
\draw [dashed, gray] (c2) -- (ax2.north east);

\end{tikzpicture}}
    \caption{Experimental measurements and fitted line utilized for the simulations. The offset created by the responser delay has been already adjusted in the raw data measurements.}
    \label{fig:uwb_meas}
\end{figure}
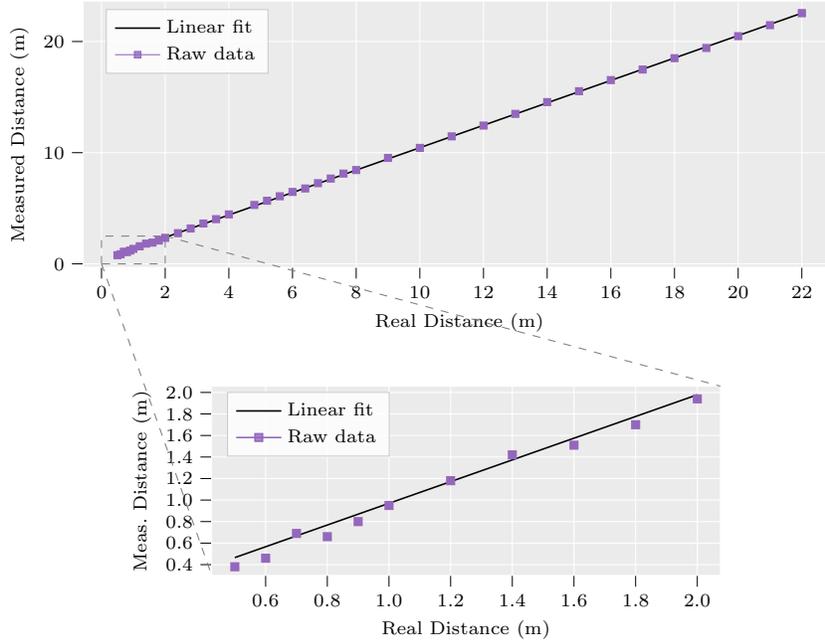

\section{Experimental Results}

\begin{figure}
    \centering 
    \begin{subfigure}[t]{0.95\textwidth}
        \centering
        \setlength{\figureheight}{0.42\textwidth}
        \setlength{\figurewidth}{\textwidth}        
        \scriptsize{
\begin{tikzpicture}

\definecolor{color0}{rgb}{0.12156862745098,0.466666666666667,0.705882352941177}
\definecolor{color1}{rgb}{0.580392156862745,0.403921568627451,0.741176470588235}
\definecolor{color2}{rgb}{0.890196078431372,0.466666666666667,0.76078431372549}

\begin{axis}[
    height=\figureheight,
    width=\figurewidth,
    axis background/.style={fill=white!92!black},
    axis line style={white},
    legend cell align={left},
    legend style={fill opacity=0.8, draw opacity=1, text opacity=1, at={(0.03,0.97)}, anchor=north west, draw=white!80!black},
    tick align=outside,
    tick pos=left,
    x grid style={white},
    xlabel={Simulation Steps},
    xmajorgrids,
    xmin=-2.7, xmax=56.7,
    xtick style={color=black},
    y grid style={white},
    ylabel={Anchor Position Error},
    ymajorgrids,
    ymin=-0.035177793909387, ymax=0.794510164560941,
    ytick style={color=black},
    ytick={-0.1,0,0.1,0.2,0.3,0.4,0.5,0.6,0.7,0.8},
    yticklabels={-0.1,0.0,0.1,0.2,0.3,0.4,0.5,0.6,0.7,0.8}
]
\addplot [semithick, color0, mark=triangle*,mark size=1.42, mark options={solid}]
table {%
0 0.206851046053415
1 0.154347876211017
2 0.138284014767492
3 0.113123663655378
4 0.0760144271867652
5 0.075662327481571
6 0.141688775379064
7 0.182556291571946
8 0.206125664822357
9 0.275595159170825
10 0.113660970340444
11 0.141235286449457
12 0.195011759825975
13 0.252279801560074
14 0.351801879406218
15 0.41620817501599
16 0.500223042120859
17 0.548803769180824
18 0.627823943337199
19 0.648014464090219
20 0.138372388806015
21 0.0517523426850758
22 0.0159560569854024
23 0.00744898228315357
24 0.0723461495568029
25 0.139800787295395
26 0.188597190064554
27 0.238781982682829
28 0.278608011874944
29 0.305041451584314
30 0.208306409841597
31 0.242796280791257
32 0.282528748531445
33 0.35528860950468
34 0.433971544243946
35 0.478900980099165
36 0.544928251178398
37 0.558881861313129
38 0.596705622196207
39 0.675565744255077
40 0.394782175216273
41 0.447757656435988
42 0.462782236071579
43 0.484156952254601
44 0.530158947480038
45 0.555764647966367
46 0.595478755032179
47 0.643191473679806
48 0.69144155125668
49 0.748097856140475
50 0.507547878861749
51 0.547302637767373
52 0.622263750696723
53 0.715590144400659
54 0.733224101333921
};
\addlegendentry{A1}
\addplot [semithick, color1, mark=square*,mark size=1.42, mark options={solid}]
table {%
0 0.0509069778778193
1 0.098271147802479
2 0.145660265457198
3 0.204441350053135
4 0.216454444382665
5 0.308202629680602
6 0.386866873937604
7 0.459834037019173
8 0.550882677426749
9 0.604901196115664
10 0.0271080991563146
11 0.0302442030871435
12 0.114609120401124
13 0.153328406245146
14 0.219488990965708
15 0.289239207894307
16 0.371773497214999
17 0.425240564863432
18 0.433337747653285
19 0.475711718096086
20 0.0975361169368866
21 0.135209728655906
22 0.213989507222138
23 0.273137096379781
24 0.316687921509998
25 0.366575082862308
26 0.406444420918692
27 0.456590087090784
28 0.480558449074374
29 0.516604665629261
30 0.226122896102159
31 0.230360590251063
32 0.243098328144232
33 0.304499175665793
34 0.369404035177165
35 0.440238464479767
36 0.491256214816991
37 0.572156123328736
38 0.608522300302378
39 0.678169066179897
40 0.29008379818301
41 0.323679595370256
42 0.378772327433608
43 0.402456199381846
44 0.432318344463195
45 0.498462818347156
46 0.553751915696583
47 0.611006327815583
48 0.659832339409925
49 0.705139756583175
50 0.633411403307854
51 0.634345860007
52 0.637601219618554
53 0.683654810842249
54 0.706766029348294
};
\addlegendentry{A2}
\addplot [semithick, color2, mark=pentagon*,mark size=1.42, mark options={solid}]
table {%
0 0.23209712489548
1 0.198659141038607
2 0.109397420058705
3 0.0675201337455393
4 0.00253529511199157
5 0.0865627721514939
6 0.16492687980779
7 0.257375714339371
8 0.28414725121383
9 0.359661978653768
10 0.243137443271113
11 0.184867640764921
12 0.154459615105374
13 0.205367272637886
14 0.202526700267848
15 0.249719306896675
16 0.275336791270801
17 0.314324020120029
18 0.381201042248557
19 0.436124150814098
20 0.17679930867802
21 0.236576039887098
22 0.299933627090918
23 0.404124253501523
24 0.467709037020239
25 0.547975636932145
26 0.583149982437528
27 0.627963058422439
28 0.709355112377255
29 0.756797075539563
30 0.357165102964099
31 0.320440940168456
32 0.322027239653284
33 0.344238940848021
34 0.383034975336112
35 0.401443584558931
36 0.439266528162909
37 0.488332025701979
38 0.548431813511927
39 0.581314135157835
40 0.0806218654181425
41 0.040772992421317
42 0.064393261119794
43 0.140038052037301
44 0.212629093570606
45 0.251352377157624
46 0.303088545002927
47 0.355050808663066
48 0.388098236025024
49 0.433421429652515
50 0.484771204866125
51 0.412955920269276
52 0.360380015009365
53 0.297944657694334
54 0.306555172527727
};
\addlegendentry{A3}
\end{axis}

\end{tikzpicture}}
        \caption{Error in the estimated position of anchors. The UWB calibration happens every ten simulation steps.}
        \label{subfig:anchor_pos_errors}
    \end{subfigure}
    \begin{subfigure}[t]{0.95\textwidth}
        \centering
        \setlength{\figureheight}{0.42\textwidth}
        \setlength{\figurewidth}{\textwidth}        
        \scriptsize{
\begin{tikzpicture}

\definecolor{color0}{rgb}{0.854901960784314,0.647058823529412,0.125490196078431}
\definecolor{color1}{rgb}{0.980392156862745,0.501960784313725,0.447058823529412}
\definecolor{color2}{rgb}{0.627450980392157,0.32156862745098,0.176470588235294}

\begin{axis}[
    height=\figureheight,
    width=\figurewidth,
    axis background/.style={fill=white!92!black},
    axis line style={white},
    legend cell align={left},
    legend style={fill opacity=0.8, draw opacity=1, text opacity=1, draw=white!80!black},
    tick align=outside,
    tick pos=left,
    x grid style={white},
    xlabel={Simulation Steps},
    xmajorgrids,
    xmin=-2.7, xmax=56.7,
    xtick style={color=black},
    y grid style={white},
    ylabel={Tag Position Error},
    ymajorgrids,
    ymin=-0.0202888014086716, ymax=0.529018648373355,
    ytick style={color=black},
    ytick={-0.1,0,0.1,0.2,0.3,0.4,0.5,0.6},
    yticklabels={-0.1,0.0,0.1,0.2,0.3,0.4,0.5,0.6}
]
\addplot [semithick, color0, mark=triangle*,mark size=2.3, mark options={solid,rotate=180}]
table {%
0 0.0462115096781704
1 0.121151376088916
2 0.14533613685534
3 0.0264101259677278
4 0.0472016350172657
5 0.0503879088301662
6 0.132636917449663
7 0.037783850181492
8 0.0264990123964002
9 0.0453668349682174
10 0.0222145503462213
11 0.0829791588585546
12 0.0753156590255472
13 0.0647648025475415
14 0.0453556634448395
15 0.0779552795052976
16 0.167458524192551
17 0.0331382278085554
18 0.133076238653665
19 0.0378133716914383
20 0.118478079083778
21 0.0642955655259322
22 0.440032431475994
23 0.161839133185005
24 0.0607483429477055
25 0.0420077559182508
26 0.0648243095430854
27 0.103285506754809
28 0.0507458816121294
29 0.0563920641931538
30 0.110144524800709
31 0.0576381234202858
32 0.176653641167803
33 0.0720444236335763
34 0.0788271178196612
35 0.155262085992848
36 0.0753849298311679
37 0.25124120071974
38 0.0894741003764399
39 0.124437502141063
40 0.102346432843062
41 0.112200137437861
42 0.1575252795674
43 0.0593391141223262
44 0.0590617774998406
45 0.26145792733683
46 0.0913906136850799
47 0.177889195953452
48 0.0388478755635993
49 0.173715151359671
50 0.121766119804244
51 0.172074255613131
52 0.188038951621647
53 0.124381574835447
54 0.137854627146669
};
\addlegendentry{Tag1}
\addplot [semithick, color1, mark=triangle*,mark size=2.3, mark options={solid}]
table {%
0 0.0100395695566515
1 0.00467971903596599
2 0.143032762946597
3 0.0133218907002928
4 0.0680066107898169
5 0.067403242730191
6 0.117417024194394
7 0.0665533569825459
8 0.0386058133681987
9 0.172104251795265
10 0.0585064405804479
11 0.19805749205475
12 0.0937066714799923
13 0.175124637467927
14 0.163214059421963
15 0.0941246888718799
16 0.0637231174748848
17 0.504050127928718
18 0.191126010166547
19 0.189575608242747
20 0.126075606469089
21 0.420858492042159
22 0.161947578368043
23 0.0770236866310498
24 0.0450036106864531
25 0.185065816525116
26 0.379530838537636
27 0.0340603086557159
28 0.0470278192451708
29 0.0661282260914493
30 0.240411425256776
31 0.168928436540779
32 0.168454244665881
33 0.0863720358341412
34 0.342096043133409
35 0.0274823730669747
36 0.131667221538174
37 0.0996217861719107
38 0.139306147543217
39 0.0525067646335392
40 0.0949998873472797
41 0.189148934044983
42 0.0975255163939895
43 0.207230800511162
44 0.210248948603282
45 0.114124790216024
46 0.05882058449724
47 0.0817760086790981
48 0.121310493747882
49 0.111435395338241
50 0.0801594213829878
51 0.120929902107156
52 0.0740784422531915
53 0.168662705499867
54 0.0997389170622713
};
\addlegendentry{Tag2}
\addplot [semithick, color2, mark=triangle*,mark size=2.3, mark options={solid,rotate=270}]
table {%
0 0.0769016933739431
1 0.0680663336857355
2 0.0534808121017522
3 0.0696926292688557
4 0.0286206249002351
5 0.118392095423411
6 0.0499520758494853
7 0.123266273705088
8 0.082020750202181
9 0.112661719757007
10 0.142129649402215
11 0.235457269384461
12 0.0914776880954947
13 0.0419784075266414
14 0.204730787950196
15 0.0942389884685361
16 0.0418750071618475
17 0.10719021241248
18 0.148488778897868
19 0.0464132559319273
20 0.0747922216776113
21 0.0693156049268132
22 0.191254201836127
23 0.128970613400911
24 0.0352866403012468
25 0.126471222833221
26 0.0928596662046327
27 0.053557396923184
28 0.0270336468341062
29 0.159232192165689
30 0.145346042038137
31 0.0553682511297993
32 0.0769090789051119
33 0.0441194267294822
34 0.0817111657861872
35 0.118827210960403
36 0.222088616774788
37 0.110002316012981
38 0.0833881441369552
39 0.1718652287899
40 0.143160109942574
41 0.062468444251477
42 0.190313045624931
43 0.0430228297504483
44 0.14192610623163
45 0.0795721667233979
46 0.0418421900272642
47 0.087889579610766
48 0.118760517942767
49 0.0319131032451077
50 0.0624004944975963
51 0.0488597785162212
52 0.128290095619435
53 0.0890911493282461
54 0.0448741616324064
};
\addlegendentry{Tag3}
\end{axis}

\end{tikzpicture}}
        \caption{Error in the estimated position of the tag during the simulation. The position of the tag is always calculated from the anchor positions based on UWB ranging.}
        \label{subfig:tag_error}
    \end{subfigure}
    \begin{subfigure}[t]{0.95\textwidth}
        \centering
        \setlength{\figureheight}{0.5\textwidth}
        \setlength{\figurewidth}{\textwidth}        
        \scriptsize{\input{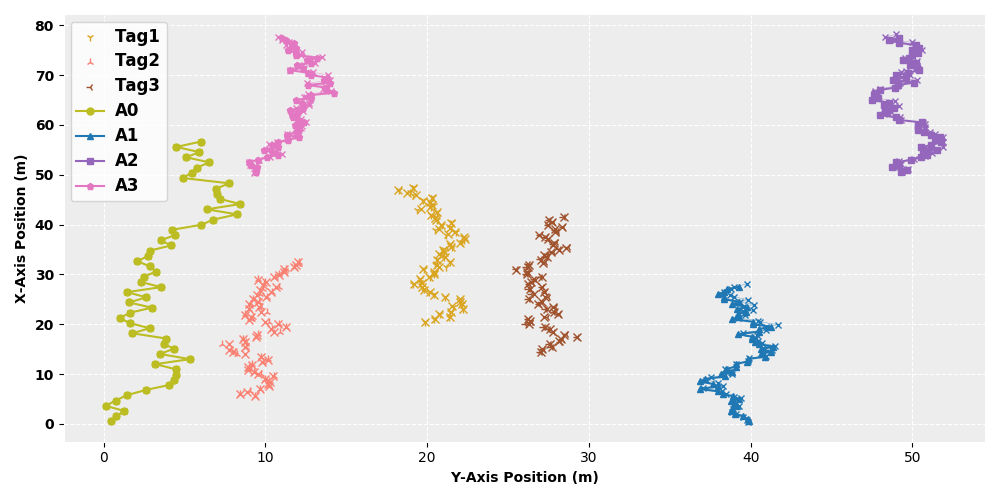}}
        \caption{Paths followed by the anchors and tags over the simulation. The paths have individual random components.}
        \label{subfig:paths}
    \end{subfigure}
    \caption{Simulation results for a system with four anchors and three tags. The figures show the error and paths over time of the anchors and tags. The position estimation of the anchors between calibrations are defined with a random error at each simulation step.}
    \label{fig:simulation}
\end{figure}
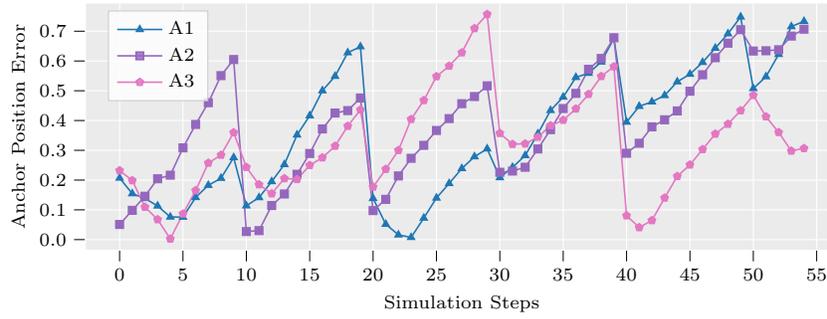
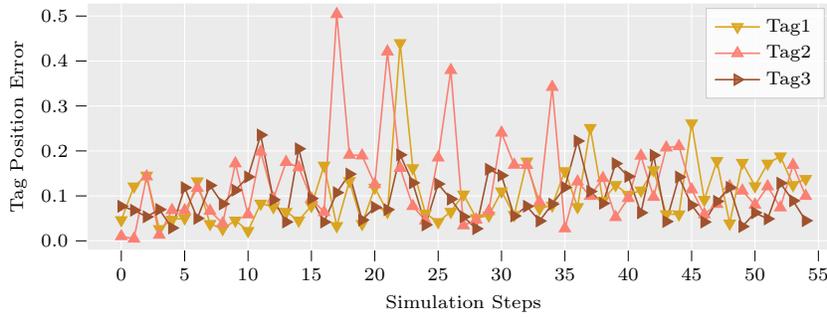
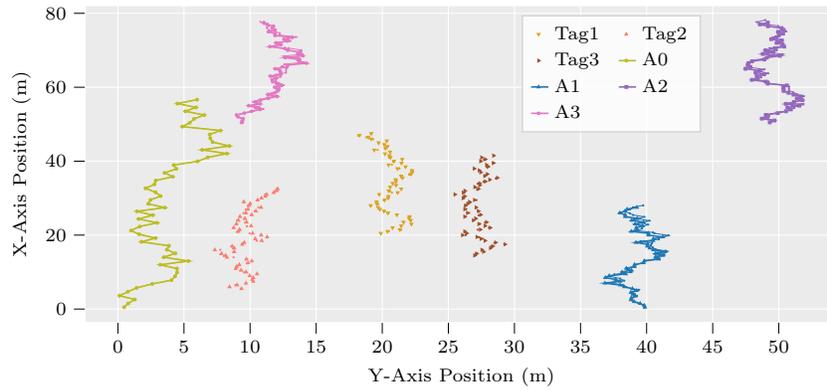

In order to test the accuracy and usability of the autocalibration algorithm, we report two different types of results. First, we have measured the accuracy of UWB ranging with the DWM1001 transceiver, and the maximum error in which our autocalibration firmware incurs has been shown in Table~\ref{tab:autopositioning}. Second, we have utilized this data to study the localization accuracy in a simulation of a mobile deployment with multiple anchors and tags.

Regarding the measurements with the DWM1001 development board, we tested our autocalibration firmware to measure its latency and accuracy. The deployed network consisted of four anchors, one of which was placed in line of sight at different distances, ranging from 0.5\,m to 22\,m. The distances measured by the UWB modules during this experiment are depicted in Fig.~\ref{fig:uwb_meas}. The results yielded from this experiment served to characterize the modules' error.

In the simulation, we have also utilized 4 anchors. A minimum of three anchors is needed, but four anchors increase the system robustness in case one of the ranging measurements fail or the error is significant~\cite{queralta2020uwb}. In addition, three tags were situated within the figure formed by the anchors to be localized. The movement of the anchors and the tags was generated following a constant direction with added random Gaussian noise. In every step, a random value in the interval $(-0.1\,m,\: +0.1\,m)$ was added to each anchor's position, representing the error of the on-board position estimation utilized between calibrations. This range of values was chosen in order to have a significant error accumulated between calibrations and test the ability of the autocalibration process to bring the error down. The anchors' calibration was performed every ten steps in the simulation. Both the calibration of the anchor positions and the positioning of the tags are done utilizing a least squares estimator, except for the initial positioning step before the movement starts. 

The results of our simulation are shown in Fig.~\ref{fig:simulation}. Subfigures~\ref{subfig:anchor_pos_errors} and~\ref{subfig:tag_error} show the error in anchors and tags positioning over 55 steps, respectively. It can be observed how calibration, performed every 10 steps, reduces significantly the anchors' positional error. The number of steps shown in this figure is reduced for visualization purposes. We have carried out multiple simulations with hundreds of steps and observed the same behavior.

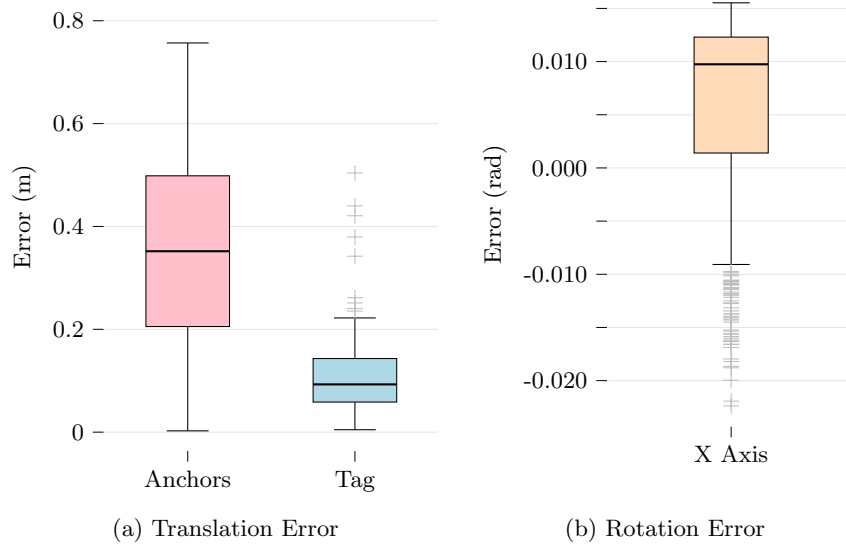
\begin{figure}
    \centering 
    \begin{subfigure}{0.52\textwidth}
        \centering
        \setlength{\figureheight}{1.23\textwidth}
        \setlength{\figurewidth}{.95\textwidth}
        \footnotesize{
\begin{tikzpicture}

\definecolor{color0}{rgb}{1,0.752941176470588,0.796078431372549}
\definecolor{color1}{rgb}{0.67843137254902,0.847058823529412,0.901960784313726}

\begin{axis}[
    height=\figureheight,
    width=\figurewidth,
    axis line style={white},
    legend style={fill opacity=0.8, draw opacity=1, text opacity=1, draw=white!80!black},
    tick align=outside,
    tick pos=left,
    x grid style={white!69.0196078431373!black},
    xmin=0.5, xmax=2.5,
    xtick style={color=black},
    xtick={1,2},
    xticklabels={Anchors, Tag},
    y grid style={white!90!black},
    ylabel={Error (m)},
    ymajorgrids,
    ymin=-0.0354955273820243, ymax=0.873857146258537,
    ytick style={color=black},
    scaled y ticks = false
]
\addplot [black, forget plot]
table {%
1 0.205367272637886
1 0.00253529511199157
};
\addplot [black, forget plot]
table {%
1 0.498462818347156
1 0.756797075539563
};
\addplot [black, forget plot]
table {%
0.875 0.00253529511199157
1.125 0.00253529511199157
};
\addplot [black, forget plot]
table {%
0.875 0.756797075539563
1.125 0.756797075539563
};
\addplot [black, forget plot]
table {%
2 0.0585064405804479
2 0.00467971903596599
};
\addplot [black, forget plot]
table {%
2 0.143160109942574
2 0.222088616774788
};
\addplot [black, forget plot]
table {%
1.875 0.00467971903596599
2.125 0.00467971903596599
};
\addplot [black, forget plot]
table {%
1.875 0.222088616774788
2.125 0.222088616774788
};
\addplot [lightgray, mark=+, mark size=3, mark options={solid}, only marks, forget plot]
table {%
2 0.235457269384461
2 0.504050127928718
2 0.420858492042159
2 0.440032431475994
2 0.379530838537636
2 0.240411425256776
2 0.342096043133409
2 0.25124120071974
2 0.26145792733683
};
\path [draw=black, fill=color0]
(axis cs:0.75,0.205367272637886)
--(axis cs:1.25,0.205367272637886)
--(axis cs:1.25,0.498462818347156)
--(axis cs:0.75,0.498462818347156)
--(axis cs:0.75,0.205367272637886)
--cycle;
\path [draw=black, fill=color1]
(axis cs:1.75,0.0585064405804479)
--(axis cs:2.25,0.0585064405804479)
--(axis cs:2.25,0.143160109942574)
--(axis cs:1.75,0.143160109942574)
--(axis cs:1.75,0.0585064405804479)
--cycle;
\addplot [thick, black, forget plot]
table {%
0.75 0.351801879406218
1.25 0.351801879406218
};
\addplot [thick, black, forget plot]
table {%
1.75 0.0928596662046327
2.25 0.0928596662046327
};
\end{axis}

\end{tikzpicture}}
        \caption{Translation Error}
        \label{subfig:boxplot_pos_error}
    \end{subfigure}
    \begin{subfigure}{0.42\textwidth}
        \centering
        \setlength{\figureheight}{1.46\textwidth}
        \setlength{\figurewidth}{.95\textwidth}
        \footnotesize{
\begin{tikzpicture}

\definecolor{color0}{rgb}{1,0.854901960784314,0.725490196078431}

\begin{axis}[
    height=\figureheight,
    width=\figurewidth,
    axis line style={white},
    legend style={fill opacity=0.8, draw opacity=1, text opacity=1, draw=white!80!black},
    tick align=outside,
    tick pos=left,
    x grid style={white!69.0196078431373!black},
    xmin=0.5, xmax=1.5,
    xtick style={color=black},
    xtick={1},
    xticklabels={X Axis},
    y grid style={white!90!black},
    ylabel={Error (rad)},
    ymajorgrids,
    ymin=-0.0292727499582575, ymax=0.01242174334837,
    ytick style={color=black},
    ytick={-0.03,-0.025,-0.02,-0.015,-0.01,-0.005,0,0.005,0.01,0.015,0.02},
    yticklabels={-0.03,-0.020,,-0.010,,0.000,,0.010,,0.02},
    scaled y ticks = false
]
\addplot [black, forget plot]
table {%
1 -0.00359699675459502
1 -0.0140767302515332
};
\addplot [black, forget plot]
table {%
1 0.00730223271647834
1 0.0105265391071596
};
\addplot [black, forget plot]
table {%
0.925 -0.0140767302515332
1.075 -0.0140767302515332
};
\addplot [black, forget plot]
table {%
0.925 0.0105265391071596
1.075 0.0105265391071596
};
\addplot [lightgray, mark=+, mark size=3, mark options={solid}, only marks, forget plot]
table {%
1 -0.0150421742256631
1 -0.016809442004059
1 -0.0193063908426505
1 -0.0215677270037755
1 -0.0160188093747541
1 -0.015875493484623
1 -0.0169342529359214
1 -0.0192195623597041
1 -0.0147379919182258
1 -0.0155458255289915
1 -0.0158992156790794
1 -0.0181529679505178
1 -0.0203486352547843
1 -0.0210124599701296
1 -0.0216076106261264
1 -0.0163443746834357
1 -0.0159641729938509
1 -0.0171661350268387
1 -0.0187300507978896
1 -0.0195009401517732
1 -0.0211951913947381
1 -0.0236536474431684
1 -0.0249640055816318
1 -0.0269298344979145
1 -0.0273775457170472
1 -0.0151512224096677
1 -0.0155865752018858
1 -0.0158768381346563
1 -0.0169775486431626
1 -0.0184446065423707
1 -0.0190808389354122
1 -0.0205895938008828
1 -0.0213212549559375
1 -0.0229558704896948
1 -0.0156948789086733
1 -0.0164330083433135
1 -0.0176865251719761
1 -0.0189619682467708
1 -0.0202414649127034
1 -0.021266985601007
1 -0.023198624563701
1 -0.0237901304775082
1 -0.0148548247799872
1 -0.015676777543178
1 -0.0162458457228077
1 -0.017498370207346
1 -0.0157226113635562
1 -0.0166671681056447
1 -0.0177533892880117
1 -0.0187342877820362
1 -0.0206371314812338
1 -0.0208474156656629
1 -0.021884497434866
};
\path [draw=black, fill=color0]
(axis cs:0.85,-0.00359699675459502)
--(axis cs:1.15,-0.00359699675459502)
--(axis cs:1.15,0.00730223271647834)
--(axis cs:0.85,0.00730223271647834)
--(axis cs:0.85,-0.00359699675459502)
--cycle;
\addplot [thick, black, forget plot]
table {%
0.85 0.0047463816768315
1.15 0.0047463816768315
};
\end{axis}

\end{tikzpicture}}
        \caption{Rotation Error}
        \label{subfig:boxplot_rot_error}
    \end{subfigure}
    \caption{Translation and rotation error distribution for the anchors and tags. The rotation error refers only to the X-axis, defined as the direction between the origin anchor and the first one in the counter-clockwise direction.}
    \label{fig:boxplots}
\end{figure}

Finally, Fig.~\ref{fig:boxplots} shows the distribution of translation and rotation errors. The translation error was calculated for both anchors and tags and is illustrated in subfigure~\ref{subfig:boxplot_pos_error}. The rotation error in subfigure~\ref{subfig:boxplot_rot_error} shows the error in the angle calculated between the x-axis and the line crossing the origin and Anchor 1. Note that Anchor 1 does not necessarily lie in the x-axis after the movement starts. In cases where the distance between these two anchors is enough this error is small. Therefore, the assumption that Anchor 1 defines the x-axis is only needed before the movement of the anchors starts.

\section{Conclusion and Future Work}


Motivated by the limitation on the applicability of UWB-based localization systems on dynamic scenarios, we have presented a mobile UWB-localization system with built-in autocalibration that can be deployed within a multi-robot system. The UWB anchors can be placed on mobile ground vehicles to support, for instance, the operation of UAVs and other robots in GNSS-denied environments. The key advantage of the proposed system is the periodic built-in self autocalibration of anchor positions. This allows for the localization error to stay within a certain tolerance even if the anchors are moving.

In future work, we will experiment with real multi-robot systems and provide a more exhaustive analysis of the usability of the proposed system in complex scenarios. We will also extend the calibration and localization approaches modelling the robots' dynamics and their odometry algorithms.


\bibliographystyle{unsrt}
\bibliography{main.bib}

\end{document}